\documentclass[twocolumn]{article}

\usepackage{ijcai24}

\usepackage{times}
\usepackage{soul}
\usepackage{url}
\usepackage[hidelinks]{hyperref}
\usepackage[utf8]{inputenc}
\usepackage[small]{caption}
\usepackage[pdftex]{graphicx}
\usepackage{amsmath}
\usepackage{amsthm}
\usepackage{booktabs}
\usepackage{algorithm}
\usepackage{algorithmic}
\usepackage[switch]{lineno} 
\usepackage{lipsum}

\usepackage{amssymb}
\usepackage{amsthm}

\usepackage{adjustbox}
\usepackage{tikz}
\usepackage{multirow}
\usepackage{xspace}
\usepackage{color}
\usepackage{caption}
\usepackage{subcaption}

\newcommand{\MethodName}{\textsc{ARU}\xspace}

\title{Attack and Reset for Unlearning: Exploiting Adversarial Noise toward Machine Unlearning through Parameter Re-initialization}

\author{
Yoonhwa Jung
\And
Ikhyun Cho\And
Shun-Hsiang Hsu\And 
Julia Hockenmaier\\
\affiliations
University of Illinois at Urbana-Champaign
\emails
\{yoonhwa2, ihcho2, hsus2, juliahmr\}@illinois.edu
}

\begin{document}

\maketitle
\begin{abstract}
   With growing concerns surrounding privacy and regulatory compliance, the concept of machine unlearning has gained prominence, aiming to selectively forget or erase specific learned information from a trained model. In response to this critical need, we introduce a novel approach called Attack-and-Reset for Unlearning (\MethodName). This algorithm leverages meticulously crafted adversarial noise to generate a parameter mask, effectively resetting certain parameters and rendering them unlearnable. \MethodName\footnote{The code is publicly available at \url{tbd}} outperforms current state-of-the-art results on two facial machine-unlearning benchmark datasets, MUFAC and MUCAC. In particular, we present the steps involved in attacking and masking that strategically filter and re-initialize network parameters biased towards the forget set. Our work represents a significant advancement in rendering data unexploitable to deep learning models through parameter re-initialization, achieved by harnessing adversarial noise to craft a mask.

\end{abstract}

\section{Introduction}
The growing emphasis on regulating the outputs of deep neural networks, fueled by concerns surrounding privacy and regulatory compliance, highlights the imperative for robust control mechanisms over these models \cite{xu2023machine,choi2023towards}. Furthermore, individuals now possess the right to demand the removal of their personal data from trained models under the \textit{Right to Be Forgotten} \cite{voigt2017eu,cali}. In response to this attention, the concept of machine unlearning has emerged, aiming to forget specific learned information or eliminate the impact of undesired data subsets from a trained model. 

A simplistic approach involves retraining the model entirely from scratch for each unlearning request, incurring substantial time and resource costs to entirely retrain the model--especially for large deep learning models. Moreover, if the volume of data to forget is substantial, retraining without those data may lead to a significantly degraded performance. To handle this challenge, various machine unlearning methods, exact and approximate machine unlearning, have been proposed. The approximate unlearning method relies on fine-tuning—initiating an unlearning process and subsequently fine-tuning the model using the retain data. In contrast, the exact machine unlearning method is rooted in ensemble learning, achieved by partitioning the size of data points.

However, these approaches are mainly designed for class-specific machine unlearning tasks, wherein specific classes are excluded from the original task (class-unlearning setting), potentially limiting their applicability in real-world scenarios \cite{choi2023towards}. Furthermore, identifying and leveraging the contribution of data features within deep learning models such as CNNs remains challenging due to their black-box nature and inherent lack of interpretability. To address these challenges, we introduce Attack-and-Reset for Unlearning (\MethodName), leveraging meticulously crafted sample-wise adversarial noise to generate a filter-level parameter mask for the precise selection and re-initialization of parameters.

The parameters of the original model encapsulate crucial knowledge from the entire training dataset that aids in solving the given task. However, intuitively, the model can be over-parameterized, containing overly detailed information that may lead to overfitting on specific feature patterns, hindering the forgetting of knowledge from the forget set. To effectively unlearn the model, the information of the retain and forget sets needs to be separated. To that end, this work develops in two folds: (1) adversarial noise attacking to disentangle the highly influenced features on the forget set, and (2) parameter masking to eliminate parameters associated with the forget data, thereby allowing a focused fine-tuning process solely on the retain data.

The adversarial noise is small and carefully crafted perturbations to the targeted data point, serving as a component of adversarial training to protect the model against adversarial attacks. Since the adversarial noise introduces nuanced perturbations to the model's initial state, we assume that a comparison between the raw image and the crafted noise allows for the interpretation of the contributions of each feature in forget set. This process aids in identifying parameters crucial for the forget data. To selectively erase sensitive features from the forget set while retaining those from the retain set, we employ the adversarial noise attack to disentangle the knowledge among features, revealing the important patterns in the forget set. Then, the parameter mask is generated to reflect which filters the features contribute to each convolutional filter of the original model.

In this paper, we specifically focus on unlearning specific instances that contain personal privacy (identity) while maintaining the original task of a given model. To the best of our knowledge, our approach is the first work to utilize adversarial noise attacks to select model parameters to be re-initialized. We validate the effectiveness of our method through comprehensive experiments on two benchmark datasets, Machine Unlearning for Facial Age Classifier (MUFAC) and Machine Unlearning for Celebrity Attribute Classifier (MUCAC). We summarize our contributions as follows:
\begin{itemize}
    \item We introduce a novel masking strategy, \MethodName, addressing the limitations of existing unlearning algorithms by selecting the highly affected convolutional filters toward forget set, achieving state-of-the-art performance.
    \item To the best of our knowledge, \MethodName is the first parameter re-initialization method for machine unlearning tasks by utilizing adversarial noise to disentangle features/patterns about data points to be forgotten.
    \item We present a thorough evaluation on two benchmark datasets, MUFAC and MUCAC, demonstrating the effectiveness and applicability of \MethodName in unlearning specific instances containing personal privacy while preserving the original task of the model.
    \item We also conduct extensive experiments to advance the understanding and application of machine unlearning, providing a robust, effective, and efficient solution for targeted data removal in deep learning models.
\end{itemize}

\label{010intro}

\section{Machine unlearning}

We briefly review the existing machine unlearning methods for deep neural networks in two folds: 1) exact machine unlearning and 2) approximate machine unlearning. 

\subsection{Exact Machine Unlearning}

Exact unlearning methods entail exact retraining but selectively operate on a subset of the dataset to avoid comprehensive retraining. SISA \cite{sisa} introduces a comprehensive exact unlearning strategy, dividing the training data into disjoint shards. Each shard is individually trained and preserved as a constituent model during the training phase. During unlearning, only the constituent model corresponding to the shard containing the data to be removed undergoes retraining, and the results are aggregated with other models. Retraining commences from the last parameter state saved before incorporating the slice containing the data point to be unlearned. \citeauthor{gupta2021adaptive} \shortcite{gupta2021adaptive} present adaptive data deletion mechanisms and \cite{yan2022arcane} utilize data preprocessing methods on top of a variant of the SISA framework. This strategy, however, demands high storage costs for the storage of multiple networks and gradient snapshots \cite{Tarun_2023}. Additionally, a reduction in shard size leads to a considerable degradation in predictive performance \cite{aldaghri2021coded}. Therefore, we focus on the approximate machine unlearning in this paper.

\subsection{Approximate Machine Unlearning}

The core concept of approximate unlearning approaches revolves around estimating the contribution of target data points to the model and adjusting the parameters for unlearning. This broad category can be roughly divided into three distinct approaches: 1) data poisoning, 2) knowledge transfer, 3) parameter manipulation, and 4) parameter masking.

\paragraph{Data Poisoning}

For machine unlearning, a data poisoning attack manipulates the training data provided to a pre-trained machine learning model, resulting in subtle changes to the model's predictions. Compared to randomly generated noise, \cite{huang2021unlearnable} utilizes an error-minimizing class-wise adversarial noise to render training examples to be forgotten for deep learning models. \citeauthor{di2022hidden} \shortcite{di2022hidden} attacks both forget set and a subset of retrain set, by adding small perturbations on the target with respect to the adversarial label. UNSIR \cite{Tarun_2023} learns an error-maximizing noise on the specific class(es), corrupting weights related to the targeted class(es) in the network. Subsequent fine-tuning on the retain set is undertaken to rectify the perturbed weights. However, this method does not identify and leverage the contributions of individual data points and features, limiting its support for unlearning a random cohort of data; it solely redirects the target forget classes into alternative classes.

\paragraph{Knowledge Transfer}
Despite the black-box nature of deep learning models, knowledge transfer methods in machine unlearning aim to selectively transfer relevant knowledge from the original model to another, minimizing the forget set's impact on model weights. Using the teacher-student framework, as in knowledge distillation, \citeauthor{chundawat2023can} \shortcite{chundawat2023can} introduces competent and incompetent teachers, where the competent teacher mirrors the original fully trained model, and the incompetent teacher is randomly initialized. Similarly, \citeauthor{zhang2023machine} \shortcite{zhang2023machine} eliminates forget set knowledge using a stochastic initialization model as a student model. Unlike joint training, they use a ``bad teacher" to erase the forget set's impact. However, maximizing the distance between the student and teacher on the forget set degrades retain set performance. Addressing this, \citeauthor{kurmanji2023scrub} \shortcite{kurmanji2023scrub} proposes SCRUB to keep the student close to the teacher on the retain set while distancing from the teacher on the forget set.

\paragraph{Parameter Manipulation}

In the realm of machine unlearning, parameter manipulation is commonly employed, leveraging the notion that weights contain information about specific subsets of the training data. \citeauthor{guo2019certified} \shortcite{guo2019certified} introduce a Gaussian noise vector and apply a Newton step on the model parameters to diminish the influence of forgettable data points. To mitigate the computationally expensive nature of calculating the Hessian matrix, \citeauthor{golatkar2020eternal} \shortcite{golatkar2020eternal} approximate the Hessian using the Fisher Information Matrix. Enhancements to this approach involve the use of the Neural Tangent Kernel (NTK) by \citeauthor{golatkar2020forgetting} \shortcite{golatkar2020forgetting} to address the null space of weights. However, the Hessian matrix still incurs high computational costs, contradicting our primary goal of efficient machine unlearning. Another approach is the concept of the gradient ascent, opposite to typical gradient descent, to remove the forget set's influence \cite{graves2021amnesiac,choi2023towards}. However, this gradient ascent can dominate the gradient descent loss term, leading to unstable performance across training epochs. Additionally, \citeauthor{goel2022cfk} \shortcite{goel2022cfk} proposes CF-k, where the last $k$ layers of the original model are trained only on the retain set, preventing catastrophic forgetting. Nonetheless, \citeauthor{golatkar2020eternal} \shortcite{golatkar2020eternal} find that fine-tuning on the retain set from the original model does not suffer catastrophic forgetting.

\paragraph{Parameter Masking}
A more efficient approach is a parameter masking strategy, where significant parameters related to the forget set are identified and removed before fine-tuning, enhancing the efficacy of unlearning. In the pursuit of escaping local optima by introducing proper perturbations to the initial state, \citeauthor{wang2022federated} \shortcite{wang2022federated} propose a TF-IDF-based masking method targeting the forgetting of a specific class. This method employs term-frequency inverse document-frequency scores to analogize channels to words and classes to documents in information retrieval, masking neurons with high TF-IDF scores and thereby pruning the contribution of the target class. \citeauthor{liu2023unlearning} \shortcite{liu2023unlearning} utilize Fisher information to characterize parameter contributions and implement a masking approach. Additionally, \citeauthor{lin2023erm} \shortcite{lin2023erm} introduces the Entanglement-Reduced Mask (ERM) structure as a learnable layer, which determines the relevance of feature maps to the knowledge of different classes by training a teacher model. Although recognizing the significance difference and entanglement between model parameters suggests that machine unlearning can benefit from resetting parameters solely associated with the forget set, current masking methods are predominantly tailored for the class-specific forgetting setting, wherein the objective is to fully misclassify a targeted class of forget set. Motivated by the concept of parameter masking, we propose a novel and efficient masking strategy, \MethodName, which leverages adversarial noise to remove the impact of the forget data points, while keeping the model performance on the original task.

\section{Problem Definition}
In this paper, the \textit{machine unlearning} task is to unlearn the set of training data, referred to as the "forget set". Compared to predominant class-specific machine unlearning tasks, our objective is to unlearn instances containing personal privacy information without altering the original task. This setting aligns more closely with real-world applications. Therefore, we adhere to the task-agnostic machine unlearning tasks, as introduced by \cite{choi2023towards}, utilizing new benchmark datasets for machine unlearning.

Given a fixed dataset $D$, the forget set is denoted as $D_f (D_f\subseteq D)$, the retain set as $D_r$, which is equivalent to $D \setminus D_f$, the original trained model as $\theta$, and the unlearned model as $\theta^{*}$. In the task-agnostic machine unlearning setting, our focus is on forgetting specific instances, while the unlearning process does not hamper the utility of the original model. The benchmark datasets and evaluation metrics are described in Section \ref{experi}. 

\section{Proposed Method: \MethodName}
In this section, we elucidate the \MethodName unlearning method in comprehensive detail. We provide the overall stages by which adversarial noise is strategically employed to generate a parameter mask for subsequent parameter re-initialization. This process culminates in the reset of knowledge associated with forget data, ultimately yielding a robust unlearned model.

\subsection{Motivation}

The original model is susceptible to over-parameterization, containing excessive intricacies that may result in overfitting to specific feature patterns, notably those within the forget set. This over-parameterization, if unaddressed, can impede the model's robustness and hinder its ability to generalize effectively. By recognizing the disparity in significance and the intricate interplay among model parameters, machine unlearning can benefit from resetting the over-parameterized elements specifically associated with information within the forget set. This selective re-initialization allows for the reduction of biases and mitigates the impact of the forget set on the overall model performance.

Intuitively, certain weights in deep learning models encapsulate general feature representations, aiding in the recognition of new data, while others may exhibit bias towards the forget dataset, contributing minimally to the final prediction. This imbalance impedes the fine-tuning process with the retain set from reaching an optimal state efficiently, manifesting the over-parameterization of the model concerning the forget set within a few steps. To effectively unlearn the forget data, a strategic re-initialization of specific parameters becomes imperative to alleviate biases and diminish the influence of the forget set before fine-tuning the model with the retain set. This approach significantly shortens the required fine-tuning steps to eradicate the impact of the forget set on the original model.

However, discerning knowledge specific to a set from a fully trained model and selecting biased weights presents a considerable challenge. To address this challenge, we leverage sample-wise adversarial noise to disentangle features highly influenced by comparing them with raw image features. This process aids in the meticulous generation of a parameter mask, which selectively identifies convolutional filters for precise unlearning of forgotten data through parameter re-initialization. This innovative approach enables the manipulation of the contributions of each feature in each data point, thereby enhancing model robustness and facilitating efficient machine unlearning.

\subsection{Attack and Reset to Unlearn (\MethodName)}
\label{aru}
\subsubsection{Stage I. Attack: Adversarial Noise Generation} The adversarial noise is meticulously crafted through multi-step attacks ($t$), employing adversarial gradients based on Projected Gradient Descent (PGD) \cite{madry2019deep}. Sample-wise adversarial noise ($\delta$) is generated for the forget data ($\mathcal{D}_f$) over pairs of examples $x \in \mathbb{R}^d$ and corresponding labels $y \in [k]$:
\[x'_0 = x\]
\[x'_{t+1} = Clip_{x,\epsilon} \{x'_t + \alpha sign(\nabla_x L(\theta,x'_t,y))\]
\[\delta = x' - x\]

Here, the noise $\delta$ is constrained by $||\delta|| \leq \epsilon$, and the small value of $\epsilon$ is set as a bound. $\alpha$ is the attack learning rate. More details can be found in the PCG paper \cite{madry2019deep}. As adversarial noise is carefully generated based on the original raw image features, it regionally retains the contributions of data features (See Fig. \ref{fig:noise}). We leverage this to identify the most influenced convolutional filters towards the forget set by exploiting the adversarial noise.

\subsubsection{Stage II. Reset: Parameter Selecting and Re-initialization}
The objective of parameter re-initialization is to mitigate overfitting and reset model parameters that may be biased toward the forget set. To achieve effective parameter selection, our focus centers on each convolutional filter within CNN-based networks. Initially, the raw images from the forget data ($x^{i}_f$) and their corresponding adversarial noises ($\delta^{i}$) undergo a single pass through the original model $\theta$. The gradient discrepancy for each convolutional layer weight between the noise and the raw image is then computed and average-pooled across different filter kernels.

\begin{figure*}[h!]
    \centering
    \includegraphics[width=\textwidth]{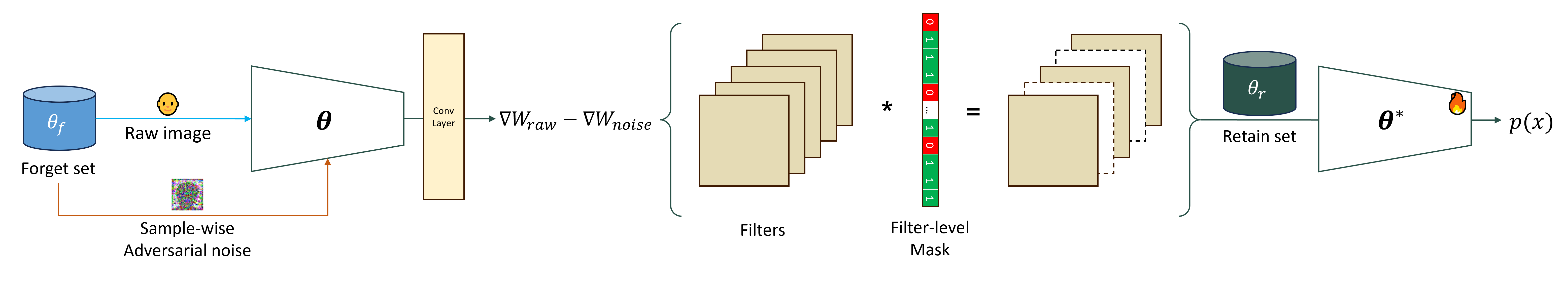}
    \caption{\MethodName: An overview of our proposed parameter masking for model re-initialization. }
    \label{fig:param_mask}
\end{figure*}

To attenuate the influence of highly affected filters in the convolutional layers, we employ a filter-level masking strategy based on the absolute discrepancy between gradients on each filter from the raw image of the forget set and its corresponding noise (Figure \ref{fig:param_mask}). Both the raw image and the adversarial noise encompass facial information like facial outline, hair, and background, as discussed in stage 1. Since different filters in convolutional layers extract different local features, the parameters may respond similarly to important patterns present in both the raw image from the forget set and its corresponding noise. Consequently, if the absolute discrepancy between gradients on each filter is small, corresponding parameters exhibit a similar reaction to identifying regional outlines and patterns. In this paper, we mask out filter kernels where this absolute discrepancy is smaller than the median value, selectively masking the 50\% most influenced convolutional filters on the forget set. This initiates the re-initialization of the layer's weight, ensuring an informed parameter reset that facilitates effective unlearning upon request and reduces overfitting toward the forget set.

\subsubsection{Stage III. Post unlearning: Finetuning}
The parameter re-initialization step may occasionally perturb the weights responsible for predicting the retain classes. To rectify this, we simply finetune on the model using the retain data, thereby ensuring a restoration of weights essential for accurate predictions.

\section{Experiments}\label{experi}
\subsection{Unlearning benchmark dataset}

Many existing machine unlearning algorithms have primarily focused on the assessment of conventional computer vision datasets such as CIFAR-10 \cite{CIFAR}, MNIST \cite{mnist}, and SVHN \cite{SVHN}. The goal of prior research based on these conventional datasets is predominantly toward unlearning specific class(es). In these settings, classification models are initially trained, and the subsequent evaluation of machine unlearning algorithms involves the intentional forgetting of specific image classes (categories) in experimental settings. Such class-unlearning scenarios may lack generalizability to real-world contexts.
To address the limitations of class unlearning and to focus on unlearning specific instances containing personal privacy information, \citeauthor{choi2023towards} \shortcite{choi2023towards} introduced two machine unlearning benchmark datasets: MUFAC and MUCAC. In this paper, we use the MUFAC benchmark dataset, and the original models for face age estimation (multi-class classification) and facial attribute prediction (multi-label classification) are provided by \citeauthor{choi2023towards} \shortcite{choi2023towards}. All images in MUFAC and MUCAC have a resolution of 128×128, depicting only the facial region. Table \ref{tab:STATS} provides an overview of the benchmark dataset statistics.

\begin{table}[ht!]
\centering
\begin{tabular}{lcc}
\toprule
 &  MUFAC   &  MUCAC  \\
\midrule
Train dataset & 10,025& 25,846\\
Test dataset & 1,539  &2,053\\
Forget dataset & 1,500 & 10,135 \\
Retain dataset & 8,525 & 15,711\\
Unseen dataset & 1,504 &2,001 \\
\bottomrule
\end{tabular}
\caption{Overall statistics of the benchmark datasets.}
\label{tab:STATS}
\end{table}

\paragraph{MUFAC (Machine Unlearning for Facial Age Classifier)}
This multi-class age classification dataset encompasses Asian facial images annotated with information regarding 8 age groups. Considering the primary objective of MUFAC is age classification, it is noteworthy that even after unlearning specific subjects (i.e., unlearning a random cohort of data), the model's overarching task remains age classification. Consequently, the aim is to sustain high accuracy for the original task even in the aftermath of the unlearning process.

\paragraph{MUCAC (Machine Unlearning for Celebrity Attribute Classifier)}

MUCAC, a multi-label facial attribute classification dataset derived from CelebA \cite{liu2015faceattributes}, comprises 30,000 images enriched with personal identity annotations for unlearning algorithms. The original problem involves multi-label classification, addressing three facial recognition tasks: male/female, old/young, and smiling/unsmiling. 

\subsection{Baseline Unlearning Methods}
\label{secbase}

\paragraph{Original Model ($\theta$)}
The benchmark datasets for machine unlearning (MUFAC and MUCAC) offer pre-trained ResNet18 models \cite{he2016deepresnet} as the original models \cite{choi2023towards}. The MUFAC original model undergoes training for 10 epochs using Stochastic Gradient Descent (SGD) with a momentum of 0.9 and a learning rate of 0.01, reduced by a factor of 10 after 10 epochs. The MUCAC original model is trained for 8 epochs using the SGD optimizer with a learning rate of 0.01.

\paragraph{Baselines}
We compare \MethodName with six approximate unlearning methods and Retrain from Scratch approach. The baselines under class-specific unlearning setting are adapted to our task-agnostic unlearning setting for a fair comparison: 1) \textbf{Retrain from Scratch:} Retrain the model from scratch only using retain data points $D_r$, 2) \textbf{Finetune:} Perform fine-tuning on the original model with the remaining data $D_r$, 3) \textbf{Negative Gradient (NegGrad):} Fine-tune the original model on the forget set $D_f$ using gradient ascent, opposite to typical gradient descent to remove the forget set's influence, 4) \textbf{Advanced Negative Gradient (AdvNegGrad) \cite{choi2023towards}:} An adjustment to NegGrad, incorporating the joint loss of fine-tuning and NegGrad in the same training batches, 5) \textbf{CF-3 \cite{goel2022cfk}:} Fine-tune only the last 3 layers of the original model on the retain set $D_r$, while freezing other layers to address catastrophic forgetting, 6) \textbf{UNSIR \cite{Tarun_2023}:} It is adjusted for MUFAC and MUCAC by synthesizing noise that maximizes the distance between the noise data and forget features \cite{choi2023towards}. Fine-tune the corrupted model on the retain set to rectify the perturbed weights, and 7) \textbf{SCRUB \cite{kurmanji2023scrub}:} As a teacher-student framework, maintain the student's proximity to the teacher on the retain set while distancing from the teacher on the forget set by maximizing the divergence for the forget data. To evaluate the general effectiveness of \MethodName in machine unlearning tasks through parameter masking, we additionally compare it with 8) \textbf{Random Masking} approach. It is noteworthy that random masking outperforms these baselines on the MUFAC and is competitive on MUCAC benchmark dataset.

\subsection{Evaluation Metrics}
Machine unlearning algorithms are typically assessed using a two-metric framework, focusing on (1) \textit{utility} and (2) \textit{forgetting} performance.

\paragraph{Utility Score}
The utility of an unlearning algorithm is evaluated by its performance on a test set, measuring the model's ability to maintain its original task performance. This aligns with the expectation that the unlearned model should continue to perform effectively in the given task post-unlearning. Following \citeauthor{choi2023towards} \shortcite{choi2023towards}, we quantify the utility score using accuracy on the test set, a score ranging from 0 to 1.

\paragraph{Forgetting quality}
To evaluate the model's forgetting capability, a fundamental aspect of machine unlearning, we employ the Membership Inference Attack (MIA) framework. In this MIA framework, a binary classifier is commonly trained to distinguish whether the model was trained on a given data point.
Using the forget dataset ($\mathcal{D}_f$) and an unseen dataset ($\mathcal{D}_u$) where the model has not been trained, we collect the model's output (cross-entropy loss) for each data point $x \in \mathcal{D}_f \cup \mathcal{D}_u$. Subsequently, a logistic regression model $\psi(\cdot)$ is trained to discern whether the input data point $x$ is from the forget or unseen dataset based on the cross-entropy loss:
\begin{gather*}
    \psi(CE(P(\hat{y}|x, \theta^*), y)) = \begin{cases}
        1 \quad \text{if } x \in \mathcal{D}_{f}\\
        0 \quad \text{if } x \in \mathcal{D}_{u}
    \end{cases}
\end{gather*}
Here, $x$ is the input, $y$ is the true label, $CE$ represents the cross-entropy loss, and $\theta^*$ denotes the unlearned model parameters. A perfect machine unlearning algorithm achieves an accuracy of 0.5, indicating indistinguishability between forget and unseen samples. The forgetting score, denoted as $|M-0.5|$, is based on the overall accuracy ($M$) of the MIA classifier.

\paragraph{NoMUS score: Normalized Machine Unlearning Score}
The NoMUS score, proposed by \citeauthor{choi2023towards} \shortcite{choi2023towards}, provides a concise single metric that aggregates utility and forgetting scores:
\begin{gather*}
    U \times \lambda + (1 - F \times 2) \times (1 - \lambda) 
\end{gather*}
Here, $U$ is the utility score, $F$ is the forget score, and $\lambda$ is a weight determining the prioritization of scores. Ranging from 0 to 1, a higher NoMUS score signifies better performance. Following \citeauthor{choi2023towards} \shortcite{choi2023towards}, we set $\lambda = 0.5$.

\subsection{Experiment details}
In pursuit of the objectives of machine unlearning, conducting an exhaustive hyperparameter search could hinder the expeditious execution of unlearning processes. Consequently, we align with default settings specified by \citeauthor{choi2023towards} \shortcite{choi2023towards} without searching hyperparameter search space. We run the experiments with ten random seeds over a uniform number of 10 epochs for all baseline models and the \MethodName and report the average results for a fair and comprehensive comparison. We used the same set of random seeds for every experiment. All experiments are conducted on a single Tesla A100 GPU. Details are provided as follows.

\paragraph{Adversarial noise generation}

The generation of meticulously crafted noises for each target data point employs the adversarial noise attack method, adopting the default hyperparameter setting from \cite{madry2019deep}. This method constitutes a robust adversary, utilizing the multi-step variant, typically involving 7 steps. For each noise per pixel, a constraint is imposed such that its Norm2 value does not exceed the fixed constraint of $\epsilon$. Specifically, $\epsilon$ is designated as $8/255$, and $\alpha$ is configured as $2/255$. The cross-entropy loss is employed to iteratively update the noise.

\paragraph{Parameter masking}

Based on our assumption, some parameter weights may be biased to the forget set and only have few contributions to the final prediction. To re-initialize the parameters, we introduce a filter-level parameter masking strategy as described in Section \ref{aru}. This approach involves the generation of a parameter mask, where 50\% of the parameters are selectively reset. While the optimal proportion of parameters to be reset may vary, we consistently set it at 50\% as the median threshold.

\paragraph{Fine-tuning}
After masking, we fine-tune parameters on the $\mathcal{D}_r$ to recover a restoration of weights essential for the final predictions. To avoid the exhaustive hyperparameter search, we fine-tune the unlearned model with the default setting by \citeauthor{choi2023towards} \shortcite{choi2023towards} both for MUFAC and MUCAC dataset; a batch size of 64, learning rate of 0.001 and a momentum of 0.9 using the SGD optimizer.

\subsection{Overall Results}
\label{overallresult}
Table \ref{tab:overall} provides a comprehensive overview of the performance of \MethodName in comparison to other competitive machine unlearning methods \cite{choi2023towards,goel2022cfk,Tarun_2023,kurmanji2023scrub}. Notably, it outperforms the previous state-of-the-art model (SCRUB) by an average of 2.85 NoMUS score, establishing a new record. Additionally, in contrast to random masking, \MethodName demonstrates the effectiveness of parameter masking across three metrics on both MUFAC and MUCAC datasets. This success solidifies \MethodName as the forefront approach in machine unlearning, utilizing strategic parameter masking through adversarial noise. 

\begin{figure}[h]
  \includegraphics[scale=0.28]{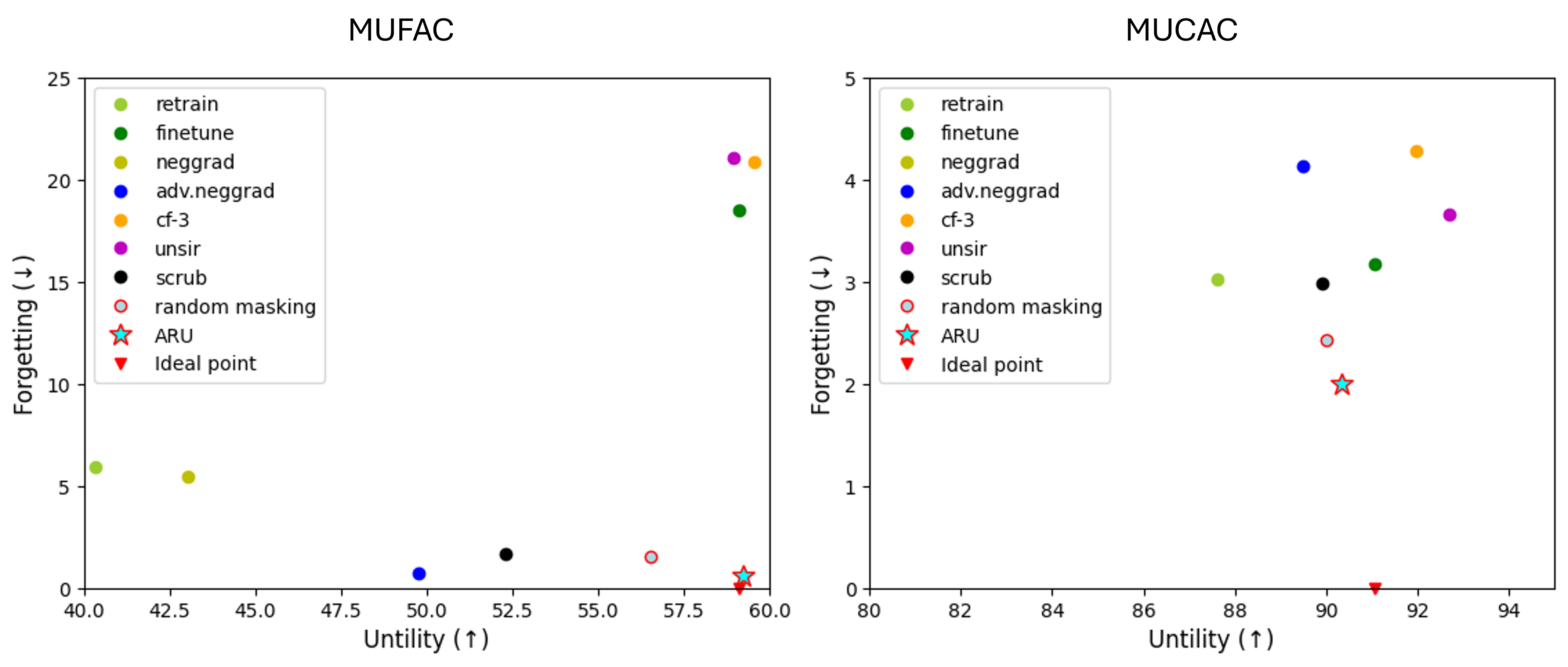}
  \centering
  \caption{Comparison of Forgetting vs. Utility among various unlearning models and \MethodName. An ideal model would exhibit a complete forgetting score (i.e., 0 Forgetting score) while maintaining a utility score equivalent to a fine-tuned model on the retain data. \MethodName shows performance closest to the ideal model.}
  \label{fig:tradeoff}
\end{figure}

In particular, \MethodName demonstrates superior forgetting ability compared to alternative unlearning methods while preserving model utility. As shown in Figure \ref{fig:tradeoff}, \MethodName sustains high accuracy for the original classification task even after the unlearning process, reaching performance closest to the ideal model. Here, the ideal model has a forgetting score of 0 and at least a utility score of a fine-tuned model on the retain data. In addition, it appears that certain machine unlearning methods, UNSIR, CF-3, and Fine-tuning, do not exhibit pronounced forgetting effects. Furthermore, given that the results are an average of ten runs, significance tests following \cite{dror2019deep} result in minimum epsilon values of 0 with $p$-value of $p$ $<$ 0.01, indicating that \MethodName's performance is ``stochastically greater'' than the baselines.

\begin{table*}[h!]
  \centering
  \begin{adjustbox}{width=1.0\textwidth,keepaspectratio}
  \begin{tabular}{l|ccc|ccc|ccc}
    \toprule
    & & MUFAC & & & MUCAC & & & Avg. & \\
    \midrule
    Model &  Utility (\%, $\uparrow$) & Forgetting (\%, $\downarrow$)& NoMUS (\%, $\uparrow$) & Utility (\%, $\uparrow$) & Forgetting (\%, $\downarrow$) & NoMUS (\%, $\uparrow$) & Utility (\%, $\uparrow$) & Forgetting (\%, $\downarrow$) & NoMUS (\%, $\uparrow$) \\
    \midrule
    Re-train & 40.32 ($\pm$2.35) & 5.95 ($\pm$0.80) & 64.19 ($\pm$0.60) & 87.62 ($\pm$3.38) & 3.03 ($\pm$1.52) & 90.77 ($\pm$1.68) & 63.97 ($\pm$2.87) & 4.49 ($\pm$1.16)& 77.48 ($\pm$1.14)\\
    Finetune & 59.13 ($\pm$0.82) & 18.52 ($\pm$0.53) & 61.05 ($\pm$0.51)  & 91.05 ($\pm$0.78) & 3.17 ($\pm$0.61) & 92.35 ($\pm$0.66)  & 75.09 ($\pm$0.80) & 10.85 ($\pm$0.57) & 76.70 ($\pm$0.59)\\
    Neg.Grad & 43.03 ($\pm$5.75) & 5.49 ($\pm$3.60) & 66.02 ($\pm$1.14) & 69.42 ($\pm$7.82) & 3.32 ($\pm$1.25) & 81.39 ($\pm$3.15) & 56.23 ($\pm$6.79) & 4.41 ($\pm$2.43) & 73.71 ($\pm$2.15)\\
    Adv.Neg.Grad
    & 49.75 ($\pm$0.79) & 0.72 ($\pm$0.41) & 74.16 ($\pm$0.54) & 89.48 ($\pm$0.56) & 4.14 ($\pm$0.82) & 90.60 ($\pm$0.74) & 69.62 ($\pm$0.68) & 2.43 ($\pm$0.62) & 82.38 ($\pm$0.64) \\
    CF-3 & 59.56 ($\pm$0.29) & 20.90 ($\pm$0.25) & 58.88 ($\pm$0.21) & 91.96 ($\pm$0.14) & 4.29 ($\pm$0.55) & 91.69 ($\pm$0.58) & 75.76 ($\pm$0.22) & 12.60 ($\pm$0.40) & 75.29 ($\pm$0.40) \\
    UNSIR  & 58.96 ($\pm$0.56) & 21.06 ($\pm$0.27) & 58.42 ($\pm$0.34) & 92.70 ($\pm$0.24) & 3.66 ($\pm$0.46) & 92.69 ($\pm$0.47) & 75.83 ($\pm$0.40) & 12.36 ($\pm$0.37) & 75.56 ($\pm$0.41)\\
    SCRUB & 52.30 ($\pm$1.10) & 1.64 ($\pm$1.21) & 74.51 ($\pm$1.00) & 89.92 ($\pm$1.69) & 2.99 ($\pm$1.08) & 91.97 ($\pm$0.70) & 71.11 ($\pm$1.40) & 2.32 ($\pm$1.15) & 83.24 ($\pm$0.85)\\ 
    \midrule
    Random.Masking & 56.54 ($\pm$1.28) & 1.53 ($\pm$1.02) & 76.73 ($\pm$0.84) &   90.01 ($\pm$1.31) &2.43 ($\pm$0.75)&92.58 ($\pm$0.55) & 73.28 ($\pm$1.30) & 1.98 ($\pm$0.89) & 84.66 ($\pm$0.70) \\
    \textbf{Ours: \MethodName} & 59.25 ($\pm$1.31) & \textbf{0.61 ($\pm$0.42)} & \textbf{79.01 ($\pm$0.49)} & 90.33 ($\pm$0.74) & \textbf{2.00 ($\pm$0.62)}& \textbf{93.17 ($\pm$0.59)} & 74.79 ($\pm$1.02) & \textbf{1.30 ($\pm$0.52)} & \textbf{86.09 ($\pm$0.54)}\\
    \bottomrule
  \end{tabular}
  \end{adjustbox}

  \caption{Overall results of \MethodName on MUFAC and MUCAC benchmarks. \MethodName sets new state-of-the-art records on all metrics on both datasets. $\uparrow/\downarrow$ indicates the metrics being larger/smaller the better, respectively. All experiments are conducted with ten random seeds and reported with the standard deviation.}
  \label{tab:overall}
\end{table*}

\section{Analysis}

\begin{table*}[ht]
\centering
\begin{small}
\adjustbox{max width=\textwidth}{
\begin{tabular}{cccccccccccccc}
\toprule
\multicolumn{1}{c}{Suppl.} &\multicolumn{1}{c}{Layer} && \multicolumn{3}{c}{MUFAC} &  & \multicolumn{3}{c}{MUCAC} &  & \multicolumn{3}{c}{Avg.} \\
\cline{1-2} \cline{3-6} \cline{8-10} \cline{12-14}
Adv.noise & Conv. && Utility ($\uparrow$) & Forgetting ($\downarrow$)& NoMUS ($\uparrow$)&& Utility ($\uparrow$) & Forgetting ($\downarrow$)& NoMUS ($\uparrow$)&& Utility ($\uparrow$) & Forgetting ($\downarrow$)& NoMUS ($\uparrow$)  \\
\midrule
\checkmark  & \checkmark & &  59.25($\pm$1.31) & 0.61($\pm$0.42) & 79.01($\pm$0.49) & & 90.33($\pm$0.74) &2.00($\pm$0.62)& 93.17($\pm$0.59) && 74.79($\pm$1.03) & 1.31($\pm$0.52) & 86.09($\pm$0.54) \\
$\triangle$ & \checkmark  & & 59.52($\pm$1.00) & 1.45($\pm$1.00) & 78.31($\pm$0.90) &&  91.04($\pm$0.60) &2.46($\pm$1.28)& 93.05($\pm$1.11) & &75.28($\pm$0.80) & 1.96($\pm$1.14) & 85.68($\pm$1.01)\\
x & \checkmark & &  59.40($\pm$1.10) & 0.89($\pm$0.70) & 78.80($\pm$0.97) & & 90.23($\pm$1.27) &2.12($\pm$1.05)& 92.99($\pm$0.84) && 74.82($\pm$1.19) & 1.51($\pm$0.88) & 85.90($\pm$0.91)\\
x & x & &  56.54($\pm$1.28) & 1.53($\pm$1.02) & 76.73($\pm$0.84) & & 90.01($\pm$1.31) &2.43($\pm$0.75)&92.58($\pm$0.55)&& 73.28($\pm$1.30) & 1.98($\pm$0.89) & 84.66($\pm$0.70)  \\
\bottomrule 
\end{tabular}
}
\end{small}
\caption{Results of the ablation study for \MethodName. The first row represents our final \MethodName model, while the last row corresponds to our baseline masking model using a 50\% random masking strategy. In the second row, a variation of our \MethodName model is presented, utilizing random noise trained solely on the forget data (i.e., error-minimizing noise on the forget set) to craft a parameter mask. The third row involves simply masking the top 50\% parameters based on the gradient on the forget set. This outcome aligns with our intuition that adversarial noise, possessing regional features such as the face outline (Figure \ref{fig:noise}), aids in selecting parameters biased toward the forget data points. All experiments were conducted with ten random seeds, and the results are reported with standard deviations.}
\label{tab:ablations}
\end{table*}

\begin{figure}[h]
  \includegraphics[scale=0.23]{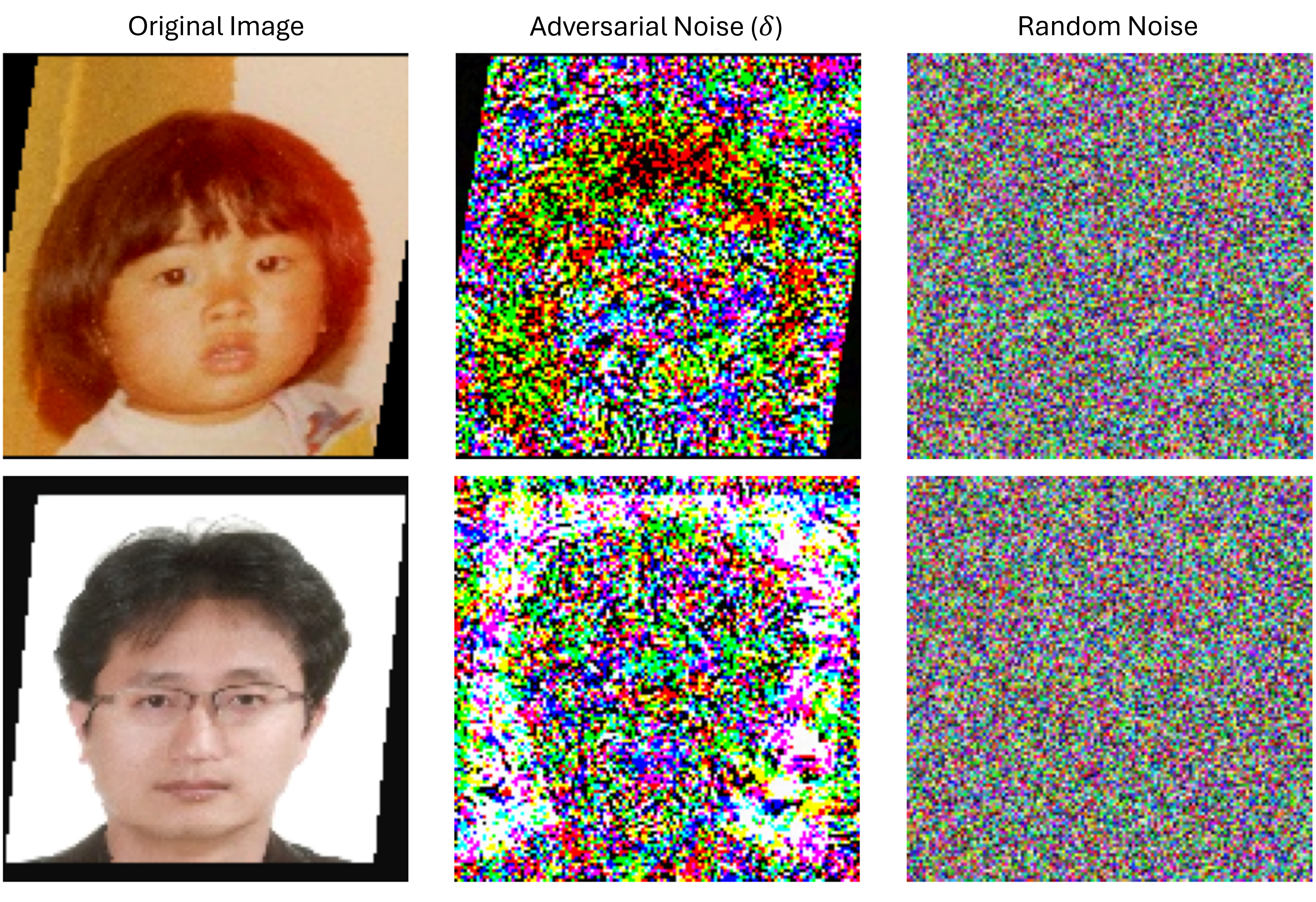}
  \centering
  \caption{Comparison between adversarial noise and random noise. Aligning our intuition, adversarial noise captures low-level information from the raw image, delineating facial and hair outlines, and background.}
  \label{fig:noise}
\end{figure}
\paragraph{Effectiveness of Adversarial Noise-based Parameter Masking}

For effective parameter selection, \MethodName is inspired by how adversarial noise is generated, reflecting the regional image feature. Table \ref{tab:ablations} presents an ablation study examining the effectiveness of leveraging adversarial noise. The comparison includes four variants: (1) \MethodName, utilizing adversarial noise to craft the filter-level parameter mask, (2) employing random noise trained on forget data labels (second row in Table~\ref{tab:ablations}), (3) simply masking the top 50\% parameters based on the gradient on the forget set to mitigate the influence of highly affected convolutional filters, and (4) randomly masking 50\% of parameters.

The results consistently demonstrate that \MethodName outperforms in all metrics across both datasets. This result aligns with our understanding of adversarial noise, which aids in selecting parameters biased toward forget data points by capturing regional features and patterns such as facial outline (Figure \ref{fig:noise}) in the forget set. Furthermore, the random noise (second row in Table \ref{tab:ablations}) proves less effective in forgetting compared to the top 50\% masking strategy (third row in Table \ref{tab:ablations}). Interestingly, random noise expedites the enhancement of the utility score (i.e., accuracy on a test set). This suggests that random noise may alleviate overfitting on the retain set, thereby improving the utility score, but it fails to effectively mitigate the influence of the forget set. In contrast, adversarial noise enables parameter re-initialization that reset forget set information, highlighting its role in the unlearning process. Random noise, lacking the ability to capture patterns from the forget set, proves less effective in parameter selection for unlearning purposes. This underscores the importance of a meticulous parameter selection strategy to facilitate the unlearning process effectively.

\begin{figure}[tb]
  \includegraphics[scale=0.25]{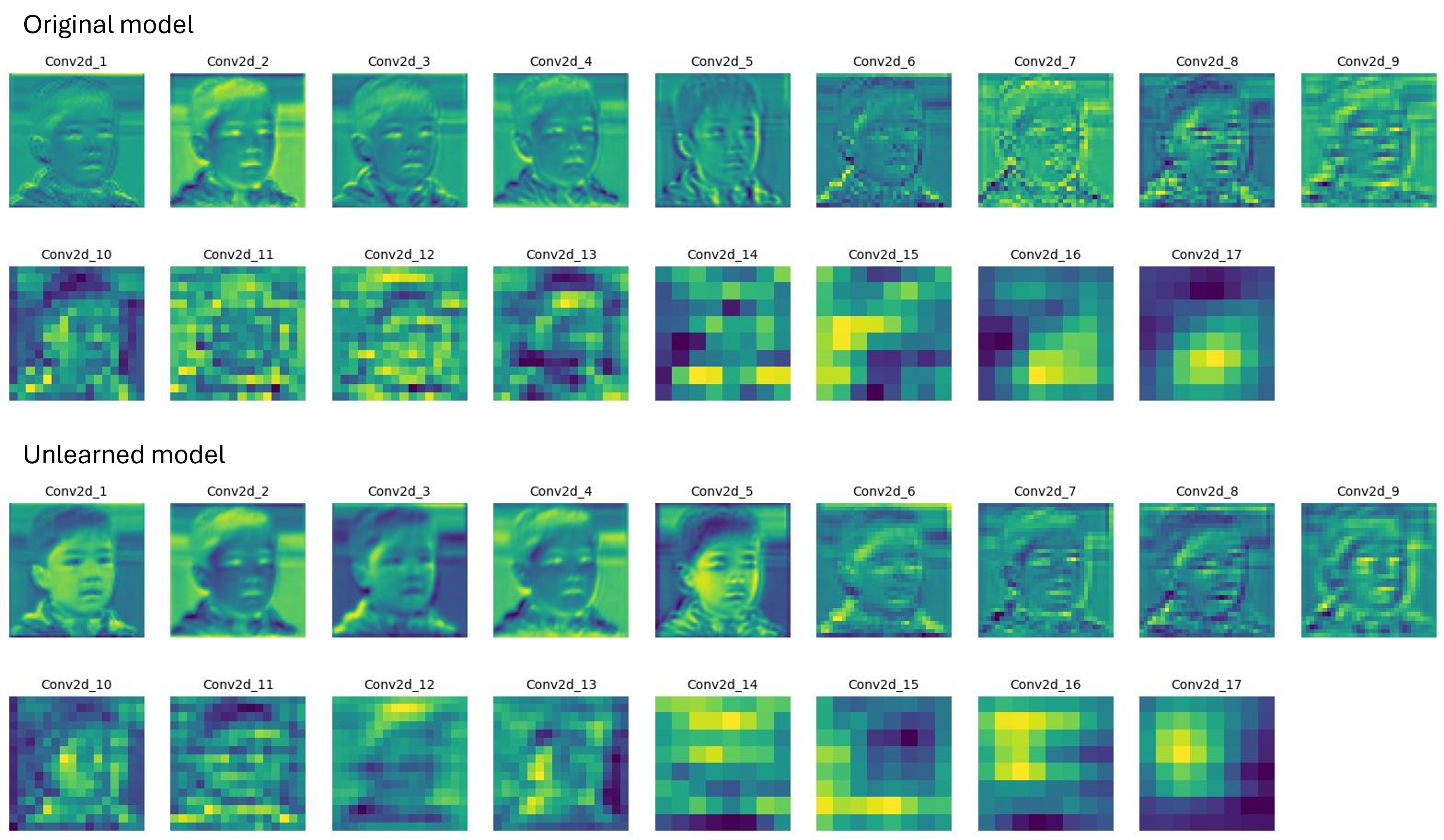}
  \centering
  \caption{Feature Map Comparison between the Original Model and Unlearned Model via \MethodName on an Unseen Data Sample}
  \label{fig:featuremap}
\end{figure}

\paragraph{Effectiveness of Parameter Re-initialization}
Given the comparatively smaller size of the MUFAC facial dataset compared to larger-scale image datasets, even smaller than MUCAC, the MUFAC model may inherently overfit to specific training set patterns, potentially limiting its ability to capture diverse local features. In Figure \ref{fig:featuremap}, we present a comparison of feature maps from different convolutional layers for an unseen MUFAC image. Despite the image being excluded from model training, the original model primarily extracts specific facial details at lower layers, while the unlearned model effectively preserves facial outlines. The strategic re-initialization of potentially redundant and biased parameters exploiting adversarial noise has led the retrained layers to achieve a more balanced extraction of low- and high-level information. Consequently, our proposed method not only mitigates the influence of the forget set but also enhances the model's robustness.

\paragraph{Efficiency Analysis}
We perform a runtime comparison between the proposed \MethodName and various competitive unlearning methods. The "retrain from scratch" approach is excluded due to its inability to achieve a utility score exceeding 55\% even after 150 epochs. Similarly, fine-tuning, CF-3, and UNSIR, as previously discussed in Section \ref{overallresult}, are omitted due to their ineffective forgetting ability. Additionally, negative gradient-based unlearning methods (Neg.Grad and Adv.Neg.Grad) are excluded as they exhibit no performance improvement over epochs, peaking within 1 to 2 epochs and subsequently showing a decline. Consequently, \MethodName demonstrates approximately 2.2 times faster performance than the previous state-of-the-art SCRUB. In summary, compared with these baseline approaches, \MethodName emerges as both more effective and efficient.

\section{Conclusion}
This paper introduces \MethodName, a novel parameter re-initialization strategy leveraging adversarial noise, to address over-parameterization and bias toward forget set for task-agnostic machine unlearning tasks. Experimental results on MUFAC and MUCAC benchmark datasets demonstrate state-of-the-art performance via effectively resetting biased parameters, rendering them unlearnable, while maintaining original task performance. Also, the strategic re-initialization with adversarial noise assists a balanced extraction of low- and high-level information from the retrained layers, diminishing the forget set's impact and enhancing model robustness. Our \MethodName represents a new paradigm for parameter re-initialization in effective, efficient, and robust machine unlearning. To the best of our knowledge, this is the first application of adversarial noise for crafting a parameter mask in this context.

\appendix

\bibliographystyle{named}
\bibliography{ijcai24}

\end{document}